\documentclass[sigconf]{acmart}
\usepackage{multirow}
\usepackage{textcomp}
\usepackage{subfig}  
\usepackage{makecell}
\usepackage{nicefrac}
\usepackage{amsthm}
\usepackage{amsfonts}
\usepackage{pifont}
\usepackage[linesnumbered,ruled,vlined]{algorithm2e}
\newcommand{\vol}{\operatorname{vol}}
\AtBeginDocument{}

\copyrightyear{2026}
\acmYear{2026}
\setcopyright{cc}
\setcctype{by}
\acmConference[CIKM '26]{Proceedings of the 35th ACM International Conference on Information and Knowledge Management}{November 07--11, 2026}{Rome, Italy}
\acmBooktitle{Proceedings of the 35th ACM International Conference on Information and Knowledge Management (CIKM '26), November 07--11, 2026, Rome, Italy}
\acmDOI{10.1145/3799682.3839943}
\acmISBN{979-8-4007-2539-5/2026/11}

\begin{document}
\title{Improved Streaming Algorithm for Balanced Hypergraph Partitioning}
\author{Zhicheng Guo}
\affiliation{
  \institution{Xidian University}
  \city{Xi'an City}
  \state{Shaanxi Province}
  \country{China}}
\email{zchguo@stu.xidian.edu.cn}
\begin{abstract}
  Balanced hypergraph partitioning is an essential task for massive hypergraph analysis in distributed computing.
  Existing methods usually use the entire hypergraph to obtain structural information before the assignment process.
  However, many applications require online partitioning, motivating the development of streaming algorithms.
  In the streaming order, vertices are assigned in real time and their ownership does not change.
  Therefore, partitioning the streamed vertices under the balance constraint is challenging.
  In this paper, we propose a method for balanced hypergraph partitioning in the streaming mode.
  By defining a modularity function under the Erdos-Renyi random hypergraph model, we build the Erdos-Renyi Hypergraph Modularity Greedy.
  We approximate the hypergraph as a uniform hypergraph and optimize the modularity function with lower computational complexity.
  Compared with other modularity functions, our modularity function is more suitable for streaming algorithms.
  Finally, we design a data structure to reduce the execution time for finding the block with minimum capacity.
  Overall, our method achieves the state-of-the-art performance when evaluated with the cut-net metric.
\end{abstract}

\ccsdesc[500]{Information systems~Data stream mining}
\ccsdesc[500]{Mathematics of computing~Hypergraphs}
\ccsdesc[500]{Mathematics of computing~Graph algorithms}

\keywords{Balanced Hypergraph Partitioning, Modularity Function, Streaming Algorithm}
\maketitle

\section{Introduction}
Different from a graph, each hyperedge in a hypergraph can connect more than two vertices \cite{gao2020hypergraph}.
When dealing with hypergraph tasks, the hypergraph is often simplified into a graph by frameworks such as Clique Expansion \cite{zien2002multilevel}, Star Expansion \cite{sun2008hypergraph}, and Line Expansion \cite{yang2022semi}.
The $k$-way Hypergraph Partitioning (HGP) problem is a generalization of the well-known graph partitioning problem \cite{papa2007hypergraph}, focusing on undirected weighted hypergraphs.
In short, this task partitions the vertex set into $k$ disjoint blocks with a balance constraint while minimizing an objective function \cite{gottesburen2024scalable}.
The common balance constraint requires that the difference in vertex capacity among the $k$ blocks be as small as possible.
In other words, the ratio between the number of vertices in the largest block and the average number of vertices is kept within a small constant $\varepsilon$.

The hypergraph partitioning task has many applications in engineering, including very-large-scale integration design \cite{karypis1997multilevel}, parallel matrix multiplication \cite{catalyurek2002hypergraph}, and combinatorial scientific computing \cite{devine2006parallel}.
In the field of Electronic Design Automation, balanced hypergraph partitioning has been widely used in various scenarios for industrial circuit design \cite{caldwell1999hypergraph}.
Hypergraph partitioning also has numerous applications in other areas, including linear algebra \cite{guo2023lahypergraph} and Boolean satisfiability \cite{duraira2004exploiting}.
To achieve efficient runtime, streaming algorithms \cite{ccatalyurek2023more} have been applied to balanced hypergraph partitioning.
Different from the multilevel scheme \cite{karypis1999multilevel}, in the streaming order, vertices are assigned in real time \cite{jiang2018highly} and their ownership does not change.
In addition, the advantages of the streaming hypergraph algorithm are faster running time and lower memory consumption \cite{mayer2018hype}.
The global structure of the hypergraph is unknown during the streaming process, so streaming methods often perform worse than multilevel methods.
On the other hand, in the streaming mode, balance among different partition blocks is considered in real time, which can prevent vertices from being assigned to the optimal block.
In this work, we aim to design a better and faster streaming algorithm for the balanced hypergraph partitioning task.

Practically efficient hypergraph partitioners from different application areas include PaToH \cite{ccatalyurek2011patoh}, hMetis \cite{karypis1999multilevel}, and KaHyPar \cite{schlag2023high}.
More recently, evolutionary algorithms such as EvoHGP \cite{andre2018memetic} and MedPart \cite{liang2024medpart} have been adopted to improve multilevel partitioning.
Beyond partition quality, more studies have considered parallel algorithms \cite{gottesburen2024scalable}, shared-memory methods \cite{gottesburen2022shared}, and acyclic hypergraph partitioning \cite{popp2021multilevel}.
In this work, we study streaming algorithms for balanced hypergraph partitioning, a line of work that emerged from MinMax \cite{alistarh2015streaming}.
The goal of streaming graph partitioning \cite{mcgregor2014graph} is to partition a vertex set into $k$ blocks while satisfying the balance constraint.
The widespread frameworks for streaming graph partitioning are Linear Deterministic Greedy \cite{stanton2014streaming} and Fennel \cite{tsourakakis2014fennel}, which are based on the Erdos-Renyi random graph model \cite{erdds1959random}.
Moreover, to improve the performance of streaming algorithms, streaming models such as the sliding window \cite{patwary2019window} and buffered streaming \cite{faraj2022buffered} have been proposed with extra computation.
For hypergraphs, the first streaming algorithm is MinMax \cite{alistarh2015streaming}, which provides theoretical guarantees.
Furthermore, the modified versions MinMax-N2P and MinMax-RE \cite{tacsyaran2021streaming} are proposed with faster running time and less memory consumption.
By extending the Fennel objective \cite{tsourakakis2014fennel} from graphs to hypergraphs, FREIGHT \cite{eyubov2025freight} achieves better performance on the cut-net metric.
Compared with the MinMax variants \cite{tacsyaran2021streaming}, the authors of FREIGHT designed a data structure for faster vertex assignment in constant time.

\paragraph{Contribution}
In this paper, we propose a method for the balanced hypergraph partitioning task. The detailed contributions are as follows:
(1) By defining a modularity function on the Erdos-Renyi random hypergraph model, we obtain a better greedy algorithm by optimizing this function.
(2) In order to reduce the running time, we design a simple and efficient data structure for the vertex assignment in the streaming order.
(3) Our method outperforms previous schemes when evaluated on the cut-net metric.

\section{Background}
\subsection{Hypergraph Notation}
Let $H = (V, E)$ be a hypergraph with vertex set $V$ and hyperedge set $E$.
There are $n=|V|$ vertices and $m=|E|$ hyperedges in the hypergraph.
Each hyperedge $e \in E$ is a subset of the vertex set $V$, in other words, $e \subseteq V$.
A vertex $v \in V$ is incident to a hyperedge $e$ if $v \in e$, and we denote by $\phi(v) = \{e \mid v \in e, e \in E \}$ the set of incident hyperedges for vertex $v$.
The size of a hyperedge $e$ is $\delta(e) = |e|$, and the degree of a vertex $v$ is $\deg(v) = |\phi(v)|$.
The volume of a vertex subset $S \subseteq V$ is defined as $\vol(S) = \sum_{v \in S} \deg(v)$.
Let $D = \{\delta(e) \mid e \in E \}$ be the set of distinct hyperedge sizes, and let $E^d \subseteq E$ be the set of hyperedges with size $d \in D$.
If all hyperedges in one hypergraph have the same size, we call it a uniform hypergraph, in other words, $|e| = d$ for all $e \in E$.
Therefore, a graph is a special case of a hypergraph when $|e| = 2$ for all hyperedges.
The $k$-way hypergraph partitioning problem splits the vertex set $V$ into $k$ non-empty disjoint blocks $\mathcal{B} = \{{B_1}, \ldots ,{B_k}\} $, such that $\bigcup _{i = 1}^k{B_i} = V$.
Given a small hyperparameter $\varepsilon$, the partition $\mathcal{B}$ is $\varepsilon$-balanced if each block $B_i$ satisfies the balance constraint:
\begin{align}
  |B_i| \le L_{\max}; \quad \quad L_{\max} = \left\lfloor (1 + \varepsilon) \frac{n}{k} \right\rfloor
  \label{equation1}
\end{align}

The set of blocks intersecting hyperedge $e$ is $\lambda(e)$, in other words, $\lambda(e) = \{ {B_i} \in \mathcal{B} \mid {B_i} \cap e \ne \emptyset \} $.
We call the $e$ an internal hyperedge if $|\lambda(e)| = 1$ and a cut hyperedge if $|\lambda(e)| > 1$.
Overall, the balanced hypergraph partitioning task seeks an $\varepsilon$-balanced $k$-way partition that minimizes an objective function on hyperedges for given $\varepsilon$ and $k$.
In this paper, we minimize the \textit{cut-net} metric on the balanced partition $\mathcal{B}$:
\begin{align}
  cut(\mathcal{B}) = \sum_{e \in E} {\mathbb{I}(|\lambda(e)| > 1)}
  \label{equation2}
\end{align}
Where $\mathbb{I}(\cdot)$ is the indicator function.
The \textit{cut-net} metric counts cut hyperedges and is used in graph partitioning.
Minimizing this objective means more internal hyperedges within each block and fewer cut hyperedges between different blocks.
In streaming mode, the vertices of a hypergraph appear one by one.
The stream elements are vertex-hyperedge pairs $(v; \phi(v))$.
For each stream element, the vertex $v$ is assigned to a block immediately.
Once a vertex is streamed, its assignment is not changed.
In addition, each stream element is forgotten after vertex $v$ is partitioned.
The partition $\mathcal{B}$ must satisfy the balance constraint in Equation \ref{equation1}.

\subsection{Streaming Hypergraph Partitioning}
Linear Deterministic Greedy \cite{stanton2014streaming} tries to place neighboring vertices in the same partition to yield fewer cut edges.
It utilizes a greedy heuristic that assigns a vertex to the partition containing most of its neighbors while respecting the balance constraint (Equation \ref{equation1}).
It selects the partition that maximizes the number of neighbors already assigned to the partition while enforcing the capacity constraint.
When a vertex $v$ arrives, MinMax \cite{alistarh2015streaming} calculates the $|E_i \cap \phi(v)|$ for each block $B_i$.
Here, $E_i$ is the hyperedge set that belongs to block $B_i$, and each hyperedge may be assigned to multiple blocks.
Each streaming vertex $v$ is assigned to the block with the largest intersection set that does not violate the balance constraint (Equation \ref{equation1}).
Thus, maximizing the number of intersecting hyperedges is a good greedy decision.
MinMax-N2P \cite{tacsyaran2021streaming} is a modified version of MinMax that stores the blocks containing each hyperedge's vertices instead of storing hyperedges per block.
Using the efficient array operations, each streamed vertex is processed in constant time.
The FREIGHT \cite{eyubov2025freight} streaming algorithm is extended from graph to hypergraph.
In this approach, each hyperedge must be assigned to a single block, and hyperedge $e$ belongs to block $\psi(e) \in [1, k]$.
More formally, the FREIGHT heuristic assigns an incoming vertex $v$ to a block $B_j$ by a modularity function.
Once a hyperedge is classified as a cut hyperedge, its contribution to the overall \textit{cut-net} metric of the hypergraph is fixed and cannot be changed.
This greedy heuristic favors the blocks with many internal hyperedges and penalizes their sizes.
Overall, the computational complexity of FREIGHT algorithm is $\mathcal{O}(\sum_{e \in E}{|e|} + n)$.

\subsection{Chung-Lu Random Hypergraph}
Under the Chung-Lu random hypergraph model \cite{lee2025survey}, hyperedges are generated according to their sizes.
By independently drawing the vertices with probability proportional to $\frac{\deg(v)}{\vol(V)}$,
the probability of drawing the vertices of a hyperedge $e$ is defined as $\mathbb{P}(e) = \prod_{v \in e} \frac{\deg(v)}{\vol(V)} $.
For block $B_i$, the expected number of hyperedges with size $d$ that are entirely contained in block $B_i$ is defined as $\mathbb{E}(B_i, d) = |E^d| \cdot (\frac{\vol(B_i)}{\vol(V)})^d$.
Therefore, the \textit{strict modularity} \cite{kaminski2019clustering} function for a partition $\mathcal{B}$ is defined as:
\begin{align}
  M(\mathcal{B}) = \frac{1}{|E|} \sum_{i = 1}^k {[|\{e \in E \mid e \subseteq B_i \}| - \sum_{d \in D}{|E^d| \cdot (\frac{\vol(B_i)}{\vol(V)})^d}]}
  \label{equation3}
\end{align}
The first term is called the \textit{edge contribution}, and the second term is called the \textit{degree tax}.
A hyperedge contributes to the first term only when it is a strictly internal hyperedge.
The \textit{degree tax} calculates the expected number of hyperedges based on the Chung-Lu hypergraph distribution.
The volume of each block determines the random probability of generating internal hyperedges.

\section{Method}
\subsection{Erdos-Renyi Hypergraph Modularity Function}
We use the connection probability between a vertex and a hyperedge to build the random hypergraph.
Let $\mathbb{P}(v \in e) = p$ be the constant probability that vertex $v$ belongs to hyperedge $e$.
We observe that the average vertex degree is $p \times m$ and the average hyperedge size is $p \times n$.
Therefore, the volume of this random hypergraph can be estimated as $p \times n \times m$.
This construction is reasonable because the volume is linearly proportional to the product of $n$ and $m$.
In this Erdos-Renyi random hypergraph, vertices are sampled with the inclusion probability $\frac{|\phi(v)|}{n \cdot m}$.
Therefore, the probability that the vertices of hyperedge $e$ are generated is denoted as $\mathbb{P}(e) = \prod_{v_i \in e} \frac{|\phi(v_i)|}{n \cdot m}$.
Let $E_i = \{e \in E \mid \psi(e) = i\}$ be the hyperedge set of block $B_i$, and each hyperedge must belong to a single block.
The expected number of hyperedges with size $d$ that are entirely contained in the block $B_i$ is $ \mathbb{E}(B_i, d) = |E^d| \cdot  \left(\frac{|B_i| \cdot |E_i|}{|V| \cdot |E|}\right)^d $.
From another perspective, the volume of block $B_i$ is estimated to be $p \times |B_i| \times |E_i|$, and the total volume of vertex set $V$ is approximated as $p \times |V| \times |E|$.
Compared with FREIGHT, the advantage of our method is that we do not need to determine the hyperparameter $p$.

The cardinality set $D$ and hyperedge set $E^d$ are unknown in the streaming process.
As a result, we approximate the ordinary hypergraph as a uniform hypergraph so that the computational complexity can be reduced to $\mathcal{O}(\sum_{e \in E}{|e|} + n)$.
The simplest method is to set $\hat{d}$ to the average size of the hyperedges that have appeared.
However, in the streaming process, previously seen hyperedges may contain vertices that have not yet appeared.
Therefore, this method causes instability in the greedy assignment.
The average hyperedge size is $\hat{d} = \frac{\vol(V)}{|E|}$.
All hyperedges in the approximated hypergraph have the same size $\hat{d}$.
Thus, the \textit{strict modularity} function in Equation \ref{equation3} can be approximated as:
\begin{align}
  \mathcal{M}(\mathcal{B}) =  \sum_{i = 1}^k {[ \frac{|\{e \in E \mid e \subseteq B_i \}|}{|E|} - { \left(\frac{|B_i| \cdot |E_i|}{|V| \cdot |E|}\right)^{\frac{\vol(V)}{|E|}}}]}
  \label{equation4}
\end{align}
We refer to Equation \ref{equation4} as \textit{approximated Erdos-Renyi modularity}.
With this modularity function, we achieve the same computational complexity as FREIGHT.

\subsection{Modularity Optimization}
In the streaming mode, each vertex-hyperedge pair (stream element) arrives one by one.
Let the stream element at time $t$ be $(v_t; \phi(v_t))$ for $1 \le t \le n$, where $v_t \in V$ and $\phi(v_t) \subseteq E$.
At time $t$, the vertices that have arrived are denoted by $V^t$, and the hyperedges that have appeared are denoted by $E^t$.
For block $B_i$ at time $t$, its vertex set is $B_i^t$, and its hyperedge set is $E_i^t$.
Since each hyperedge can only belong to one block, the set of blocks connected to vertex $v_t$ is $\{ \psi(e) | e \in \phi(v_t) \}$.
Thus, the modularity gain from Equation \ref{equation4} is calculated by:
\begin{align}
  j = \mathop{\arg\max}_{\substack{i \in \{ \psi(e) \mid e \in \phi(v_t) \} \\ |B_i^t| < L_{\max}}} \{ \Delta^e_i(v_t) + \Delta^d_i(v_t) \}
  \label{equation5}
\end{align}
The greedy search finds the maximum modularity gain over the blocks connected to vertex $v_t$.
In Equation \ref{equation5}, the first term $\Delta^e_i(v_t)$ is \textit{edge contribution gain}, and the second term $\Delta^d_i(v_t)$ is \textit{degree tax gain}.
Let $\omega(e)$ be the number of vertices in hyperedge $e$ that belong to the same block $\psi(e)$.
Thereby, the edge contribution gain is calculated by:
\begin{align}
  \Delta^e_i(v_t) = \frac{|\{ e \in \phi(v_t) \mid \omega(e) = |e| - 1, e \in E_i^t \}|}{|E^t|}
  \label{equation6}
\end{align}
When $\omega(e) = |e| - 1$, assigning vertex $v_t$ to block $i$ ensures that hyperedge $e$ is completely contained in set $B_i^t \cup \{v_t\}$.
Compared with the FREIGHT, whose modularity function is defined on the random graph, our method considers the case where all vertices of a hyperedge are entirely contained in the same block.
In Equation \ref{equation6}, a hyperedge $e$ contributes the gain only if it is strictly internal in the block $i$.
Let $\hat{d}_t = \frac{\vol(V^t)}{|E^t|}$ be the approximated hyperedge size for a uniform hypergraph.
Thereby, the degree tax gain could be computed by:
\begin{align}
  \Delta^d_i(v_t) =  \left(\frac{|B_i^t| \cdot |E_i^t|}{|V^t| \cdot |E^t|}\right)^{\hat{d}_t} - \left(\frac{(|B_i^t| + 1) \cdot |E_i^t \cup \phi(v_t)|}{|V^t| \cdot |E^t|}\right)^{\hat{d}_t}
  \label{equation7}
\end{align}
When the vertex $v_t$ is assigned to the block $i$, the block capacity increases by $1$, and its hyperedge set becomes $E_i^t \cup \phi(v_t)$.
Finally, we refer to the method in Equation \ref{equation4} and Equation \ref{equation5} as \textit{Erdos-Renyi Hypergraph Modularity Greedy} (\textsf{ERHMG}), which is designed for balanced hypergraph partitioning in the streaming mode.

First, we simply cache three variables in the streaming algorithm: $|V^t|$, $|E^t|$, and $\vol(V^t)$.
Second, there are two arrays of size $k$ in the memory: $|B_i^t|$ and $|E_i^t|$.
Third, there are two variable-size arrays in the memory: $\psi(e)$ and $\omega(e)$ for all $e \in E$.
Finally, the assignment array stores all vertices and has maximum size $n$.
Our streaming algorithm has memory consumption similar to that of FREIGHT and MinMax-N2P.
Equation \ref{equation5} has the computational complexity $\mathcal{O}(\sum_{e \in E}{|e|})$, and extra computation $\mathcal{O}(n)$ is needed to handle the case $|B_i^t| = L_{\max}$ for all candidate blocks.
Overall, the computational complexity of ERHMG is $\mathcal{O}(\sum_{e \in E}{|e|} + n)$, which is comparable to the FREIGHT.

\begin{table*}[!t]
  \caption{Comparison of the cut-net metric for different methods evaluated on the ISPD2006 Benchmark.}
  \begin{center}
    \begin{tabular}{c|c|c|c|c|c|c|c|c|c|c}
      \hline
      $k$                  & $\varepsilon$         & Method                          & newblue1        & newblue2        & newblue3        & newblue4        & newblue5        & newblue6        & newblue7         & adaptec5        \\
      \hline
      \multirow{5}{*}{128} & \multirow{5}{*}{0.02} & MinMax                          & 221204          & 293945          & 314900          & 418659          & 799410          & 828412          & 1618596          & 542646          \\
      \                    & \                     & MinMax-N2P                      & 221630          & 294037          & 315602          & 417448          & 791263          & 821130          & 1604143          & 536567          \\
      \                    & \                     & LDG                             & 222678          & 293435          & 314705          & 424297          & 819067          & 839793          & 1665980          & 545792          \\
      \                    & \                     & FREIGHT                         & 174359          & 244416          & 270071          & 351898          & 624033          & 687946          & 1337701          & 452261          \\
      \cline{3-11}
      \                    & \                     & \textbf{ERHMG}           (ours) & \textbf{164478} & \textbf{234456} & \textbf{257072} & \textbf{336125} & \textbf{578171} & \textbf{649263} & \textbf{1238699} & \textbf{420540} \\
      \hline
      \multirow{5}{*}{128} & \multirow{5}{*}{0.03} & MinMax                          & 219520          & 291042          & 311635          & 416182          & 795729          & 824653          & 1615429          & 538928          \\
      \                    & \                     & MinMax-N2P                      & 218669          & 291582          & 312814          & 414806          & 789354          & 817684          & 1601964          & 533955          \\
      \                    & \                     & LDG                             & 222678          & 293435          & 314705          & 424297          & 819067          & 839793          & 1665980          & 545792          \\
      \                    & \                     & FREIGHT                         & 174359          & 244416          & 270071          & 351898          & 624033          & 687946          & 1337701          & 452261          \\
      \cline{3-11}
      \                    & \                     & \textbf{ERHMG}           (ours) & \textbf{163727} & \textbf{233843} & \textbf{256096} & \textbf{334823} & \textbf{575495} & \textbf{646878} & \textbf{1234316} & \textbf{418869} \\
      \hline
      \multirow{5}{*}{256} & \multirow{5}{*}{0.02} & MinMax                          & 229336          & 303808          & 326437          & 429242          & 814484          & 842576          & 1635568          & 558238          \\
      \                    & \                     & MinMax-N2P                      & 230789          & 305138          & 327624          & 429075          & 806249          & 839322          & 1625751          & 555766          \\
      \                    & \                     & LDG                             & 225717          & 295282          & 318704          & 427162          & 825256          & 845729          & 1677956          & 550392          \\
      \                    & \                     & FREIGHT                         & 174852          & 245303          & 271133          & 352353          & 626752          & 690649          & 1343955          & 453637          \\
      \cline{3-11}
      \                    & \                     & \textbf{ERHMG}           (ours) & \textbf{164764} & \textbf{235018} & \textbf{258198} & \textbf{337136} & \textbf{580257} & \textbf{650035} & \textbf{1242660} & \textbf{422556} \\
      \hline
      \multirow{5}{*}{256} & \multirow{5}{*}{0.03} & MinMax                          & 225797          & 299367          & 321321          & 425411          & 808960          & 836711          & 1631998          & 552302          \\
      \                    & \                     & MinMax-N2P                      & 226321          & 300306          & 321702          & 424311          & 803256          & 832205          & 1623888          & 548280          \\
      \                    & \                     & LDG                             & 225717          & 295282          & 318704          & 427162          & 825256          & 845729          & 1677956          & 550392          \\
      \                    & \                     & FREIGHT                         & 174852          & 245303          & 271133          & 352353          & 626752          & 690649          & 1343955          & 453637          \\
      \cline{3-11}
      \                    & \                     & \textbf{ERHMG}  (ours)          & \textbf{164076} & \textbf{234452} & \textbf{257185} & \textbf{336071} & \textbf{577622} & \textbf{647567} & \textbf{1238364} & \textbf{420919} \\
      \hline
    \end{tabular}
  \end{center}
  \label{table2}
\end{table*}

\subsection{Improved Data Structure}
\label{section}
When a vertex arrives, if all blocks connected to it have reached the capacity constraint $L_{\max}$, we need to find the block with the minimum capacity.
It is necessary to design a data structure in the streaming algorithm to find this block with little computation.
MinMax-N2P defines the current minimum-capacity block as $B_{\min}$ and updates the $B_{\min}$ only when a vertex is assigned to block $B_{\min}$.
Thus, the computation cost of each update in the data structure is $\mathcal{O}(\log(k))$.
Overall, MinMax-N2P needs the extra computation $\mathcal{O}(n \cdot \log(k))$ to handle this situation.

The complex data structure of FREIGHT is implemented via two arrays and a bucket list.
The first array stores the capacity of all $k$ blocks in ascending order.
The second array maps the index $i$ of a block $B_i$ to its position in the first array.
Each bucket is associated with a block capacity and the leftmost and rightmost positions in the first array.
This data structure keeps the blocks sorted by capacity.
When a vertex has no connected blocks that satisfy the capacity constraint, this vertex is assigned to the block corresponding to the first element in the first array, which has the minimum capacity.
Thus, for each arrived vertex, the two arrays and bucket list are updated in $\mathcal{O}(1)$ time.
In short, the extra computation of data structure in FREIGHT is $\mathcal{O}(n)$.

\begin{algorithm}[!h]
  \caption{Efficient data structure for vertex assignment.}
  \label{algorithm1}
  Initialize $\texttt{Queue}[\cdot]$ as the queue array with size $L_{\max}+1$;\\
  Let $c$ be the current minimum capacity, and set $c = 0$;\\
  For each block $i$ from $1$ to $k$, set $\texttt{Queue}[0].push(i)$.\\
  \ForEach{$(v; \phi(v))$ in streaming order}
  {
    Find the candidate block set $S = \{ \psi(e) | e \in \phi(v) \}$;\\
    Seek the best block index $j$ based on Equation \ref{equation5};\\
    \If{$S = \emptyset$ or $|B_j| = L_{\max}$}
    {
      $j = \texttt{Queue}[c].front()$;\\
      \While{$c \neq |B_j|$}
      {
        $\texttt{Queue}[|B_j|].push(j)$; \\
        $\texttt{Queue}[c].pop()$;\\
        \While{$\texttt{Queue}[c].empty()$}{$c = c + 1;$}
        $j = \texttt{Queue}[c].front()$;
      }
      $\texttt{Queue}[|B_j|+1].push(j)$; \\
      $\texttt{Queue}[c].pop()$;\\
      \If{$\texttt{Queue}[c].empty()$}{$c = c + 1;$}
    }
    $B_j = B_j \cup \{v\}$;
  }
\end{algorithm}

Although the cost of updating the data structure in FREIGHT is $\mathcal{O}(1)$, its two arrays and bucket list are accessed multiple times.
When there are too many vertices, finding the best block is very time-consuming.
Thus, we design a data structure that has the same $\mathcal{O}(1)$ update time but is more efficient in practice.
First, we use the queue array $\texttt{Queue}[\cdot]$ with size $L_{\max}+1$ to store the indices of the $k$ blocks.
Second, a variable $c$ records the current minimum capacity, and its initial value is $0$.
Third, the first queue $\texttt{Queue}[0]$ is set to the range from $1$ to $k$.
Finally, the data structure for finding the minimum-capacity block is shown in Algorithm \ref{algorithm1}.
During execution, the queue array $\texttt{Queue}[\cdot]$ always maintains the same $k$ elements.
Algorithm \ref{algorithm1} reduces unnecessary computation because the queue array is updated only when no suitable block is found.
The extra computation of ERHMG is $\mathcal{O}(n)$, and it runs faster than FREIGHT and MinMax-N2P.

\begin{table*}[!t]
  \caption{The running time (second) for streaming algorithms with different data structures executed on ISPD2006 Benchmark.}
  \begin{center}
    \begin{tabular}{c|c|c|c|c|c|c|c|c|c|c}
      \hline
      $k$                  & $\varepsilon$         & Method                 & newblue1          & newblue2          & newblue3          & newblue4          & newblue5          & newblue6          & newblue7         & adaptec5          \\
      \hline
      \multirow{3}{*}{128} & \multirow{3}{*}{0.02} & MinMax-N2P             & 0.26746           & 0.372049          & 0.41452           & 0.557474          & 1.14421           & 1.47229           & 2.63953          & 0.771279          \\
      \                    & \                     & FREIGHT                & 0.182589          & 0.252773          & 0.280897          & 0.394436          & 0.815751          & 0.920045          & 2.0712           & 0.547697          \\
      \cline{3-11}
      \                    & \                     & \textbf{ERHMG}  (ours) & \textbf{0.17449}  & \textbf{0.246399} & \textbf{0.277025} & \textbf{0.375782} & \textbf{0.798038} & \textbf{0.91843}  & \textbf{2.06382} & \textbf{0.527916} \\
      \hline
      \multirow{3}{*}{128} & \multirow{3}{*}{0.03} & MinMax-N2P             & 0.264269          & 0.372244          & 0.404298          & 0.556723          & 1.25123           & 1.2563            & 2.66163          & 0.761262          \\
      \                    & \                     & FREIGHT                & 0.176584          & 0.254742          & 0.283716          & 0.388509          & 0.92126           & 0.91199           & 2.06574          & 0.55499           \\
      \cline{3-11}
      \                    & \                     & \textbf{ERHMG}  (ours) & \textbf{0.172555} & \textbf{0.251349} & \textbf{0.271746} & \textbf{0.376282} & \textbf{0.871494} & \textbf{0.900867} & \textbf{1.97415} & \textbf{0.534898} \\
      \hline
      \multirow{3}{*}{256} & \multirow{3}{*}{0.02} & MinMax-N2P             & 0.274295          & 0.544766          & 0.439675          & 0.609976          & 1.21933           & 1.29214           & 2.72455          & 0.825117          \\
      \                    & \                     & FREIGHT                & 0.179409          & 0.289805          & 0.322544          & 0.41705           & 0.875628          & 0.912044          & 2.02725          & 0.585969          \\
      \cline{3-11}
      \                    & \                     & \textbf{ERHMG}  (ours) & \textbf{0.175219} & \textbf{0.278556} & \textbf{0.300232} & \textbf{0.390644} & \textbf{0.846244} & \textbf{0.883746} & \textbf{2.02246} & \textbf{0.534297} \\
      \hline
      \multirow{3}{*}{256} & \multirow{3}{*}{0.03} & MinMax-N2P             & 0.282629          & 0.409018          & 0.44683           & 0.614593          & 1.2612            & 1.24457           & 2.72587          & 0.832845          \\
      \                    & \                     & FREIGHT                & 0.185196          & 0.279647          & 0.300234          & 0.418296          & 0.906391          & 0.876033          & 2.06947          & 0.605885          \\
      \cline{3-11}
      \                    & \                     & \textbf{ERHMG}  (ours) & \textbf{0.181631} & \textbf{0.255096} & \textbf{0.28763}  & \textbf{0.408647} & \textbf{0.899476} & \textbf{0.843226} & \textbf{2.02666} & \textbf{0.590316} \\
      \hline
    \end{tabular}
  \end{center}
  \label{table4}
\end{table*}

\section{Experiment}
\subsection{Datasets}
To evaluate the performance of our algorithm, we adopt two benchmark datasets.
The first hypergraph dataset is converted from the ISPD 2006 Benchmark Suite \cite{ISPD2006}, which is used for circuit placement contests \cite{markov2012progress}.
The structural information of the 8 hypergraphs in the ISPD2006 Benchmark is displayed in Table \ref{table1}.
\begin{table}[!h]
  \caption{Statistics of 8 hypergraphs in ISPD2006 Benchmark.}
  \begin{center}
    \setlength{\tabcolsep}{2pt}
    \begin{tabular}{c|c|c|c|c}
      \hline
      Name     & \makecell{$|V|$} & \makecell{$|E|$} & \makecell{Average          \\ hyperedge size}  & \makecell{Average \\ vertex degree} \\
      \hline
      newblue1 & 330474           & 329688           & 3.5489            & 3.5405 \\
      newblue2 & 441509           & 452200           & 3.6058            & 3.6931 \\
      newblue3 & 494010           & 530830           & 3.3139            & 3.5609 \\
      newblue4 & 646132           & 629778           & 3.731             & 3.6366 \\
      newblue5 & 1233057          & 1249244          & 3.7316            & 3.7805 \\
      newblue6 & 1255038          & 1274355          & 3.9183            & 3.9786 \\
      newblue7 & 2507953          & 2606236          & 3.6866            & 3.831  \\
      adaptec5 & 843127           & 858570           & 3.8461            & 3.9165 \\
      \hline
    \end{tabular}
  \end{center}
  \label{table1}
\end{table}

The second hypergraph dataset is transformed from the MLCAD 2023 FPGA Macro Placement Contest \cite{MLCAD2023}, whose benchmark suite has 140 hypergraphs.
The number of vertices in the MLCAD2023 Benchmark ranges from 561,064 to 707,673, and the number of hyperedges ranges from 552,386 to 697,920.
These 140 hypergraphs have the same maximum vertex degree of 57 and a maximum hyperedge size of 654.
Obviously, the structural differences between each hypergraph in this dataset are very small, suitable for testing runtimes.
In the streaming mode of hypergraph partitioning, we stream the hypergraphs with the natural given order of the vertices.

\subsection{Implementation Details}
We implement the algorithm in \textit{C++20} and compile the code with \textit{G++ 11.4}.
All the experiments are run on an Intel Xeon CPU with 64 GB memory.
We compare our method ERHMG with four previous methods:
MinMax \cite{alistarh2015streaming}, MinMax-N2P \cite{tacsyaran2021streaming}, LDG \cite{stanton2014streaming}, and FREIGHT \cite{eyubov2025freight}.
Finally, we evaluate the \textit{cut-net} metric (lower is better) for balanced hypergraph partitioning in Equation \ref{equation2}.
The tested settings combine the numbers of partition blocks $k \in \{128, 256\}$ and balance constraints $\varepsilon \in \{0.02, 0.03\}$.

\subsection{Hypergraph Partitioning Results}
We obtain the experimental results of five different methods on the ISPD2006 Benchmark with 8 hypergraphs, as shown in Table \ref{table2}.
In addition, the evaluation results on the MLCAD2023 Benchmark (averaged over 140 hypergraphs) are shown in Table \ref{table3}.
Compared with other schemes, our scheme achieves better cut-net results on both datasets.
MinMax, MinMax-N2P, and LDG have the similar hypergraph partitioning performance.
However, FREIGHT and ERHMG have substantially lower cut-net values than these three methods.
In summary, our method achieves the state-of-the-art performance on ISPD2006 Benchmark and MLCAD2023 Benchmark with the same or lower computational complexity than previous methods.
Of course, the smaller the balance constraint $\varepsilon$, the more difficult it is to partition the hypergraph.
Likewise, the larger the number of blocks, the smaller the block-size penalty in the streaming algorithm.
According to  Table \ref{table2} and Table \ref{table3}, our algorithm is robust to different partitioning hyperparameter settings.

\begin{table}[!h]
  \caption{Average \textit{cut-net} metric for different methods evaluated on the MLCAD2023 Benchmark with 140 hypergraphs.}
  \begin{center}
    \begin{tabular}{c|c|c|c|c}
      \hline
      \multirow{2}{*}{Method} & \multicolumn{2}{c|}{$k = 128$} & \multicolumn{2}{c}{$k = 256$}                                               \\
      \cline{2-5}
      \                       & $\varepsilon = 0.02$           & $\varepsilon = 0.03$          & $\varepsilon = 0.02$ & $\varepsilon = 0.03$ \\
      \hline
      MinMax                  & 375201                         & 373746                        & 382500               & 378732               \\
      MinMax-N2P              & 374944                         & 373447                        & 382322               & 378720               \\
      LDG                     & 383172                         & 383172                        & 384946               & 384946               \\
      FREIGHT                 & 343385                         & 343385                        & 344560               & 344560               \\
      \hline
      \textbf{ERHMG} (ours)   & \textbf{339084}                & \textbf{337818}               & \textbf{339850}      & \textbf{338575}      \\
      \hline
    \end{tabular}
  \end{center}
  \label{table3}
\end{table}

To test the efficiency of our proposed data structure,
we have evaluated the algorithm running time on ISPD2006 Benchmark (8 hypergraphs) and MLCAD2023 Benchmark (140 hypergraphs).
We replace the data structure in the ERHMG algorithm with the data structures from MinMax-N2P and FREIGHT, which are described in Section \ref{section}.
Table \ref{table4} and Table \ref{table5} show the difference in running time for these three data structures.
Even though FREIGHT and ERHMG have the same computational complexity, ERHMG has a slightly lower execution time.
The simple data structure in MinMax-N2P has a much longer runtime and is not suitable for streaming algorithms.
In short, running time is crucial for balanced hypergraph partitioning in streaming mode.
\begin{table}[!h]
  \caption{Average running time (second) for streaming algorithms with different data structures executed on MLCAD2023 Benchmark (140 hypergraphs).}
  \begin{center}
    \begin{tabular}{c|c|c|c|c}
      \hline
      $k$ & $\varepsilon$ & MinMax-N2P & FREIGHT  & \textbf{ERHMG}    \\
      \hline
      128 & 0.02          & 0.557854   & 0.390243 & \textbf{0.384195} \\
      128 & 0.03          & 0.562774   & 0.393786 & \textbf{0.385878} \\
      256 & 0.02          & 0.579528   & 0.391928 & \textbf{0.384821} \\
      256 & 0.03          & 0.581045   & 0.394432 & \textbf{0.386922} \\
      \hline
    \end{tabular}
  \end{center}
  \label{table5}
\end{table}

\section{Conclusion}
In this paper, we have presented a method to improve the quality of balanced hypergraph partitioning in streaming mode.
Our method, \textit{Erdos-Renyi Hypergraph Modularity Greedy}, is based on the Erdos-Renyi random hypergraph model and optimizes the cut-net metric.
Furthermore, we design an efficient data structure to reduce the running time of the streaming algorithm.
In particular, our scheme is well suited to the balanced hypergraph partitioning task.
Moreover, our method outperforms previous methods in both cut-net metric and runtime.
To summarize, we expect that the proposed modeling approach can be applied to other tasks on the hypergraph.

\section*{GenAI Usage Disclosure}
The authors not use the Generative AI tools to assist with grammar and lexical polishing, equation and symbol consistency checks, and drafting this disclosure.
The authors reviewed and edited all changes and take full responsibility for the paper's content, including the methods, code, data, experimental results, and references.

\bibliographystyle{ACM-Reference-Format}
\balance
\bibliography{sample-base}


\begin{thebibliography}{36}


\ifx \showCODEN    \undefined \def \showCODEN     #1{\unskip}     \fi
\ifx \showISBNx    \undefined \def \showISBNx     #1{\unskip}     \fi
\ifx \showISBNxiii \undefined \def \showISBNxiii  #1{\unskip}     \fi
\ifx \showISSN     \undefined \def \showISSN      #1{\unskip}     \fi
\ifx \showLCCN     \undefined \def \showLCCN      #1{\unskip}     \fi
\ifx \shownote     \undefined \def \shownote      #1{#1}          \fi
\ifx \showarticletitle \undefined \def \showarticletitle #1{#1}   \fi
\ifx \showURL      \undefined \def \showURL       {\relax}        \fi
\providecommand\bibfield[2]{#2}
\providecommand\bibinfo[2]{#2}
\providecommand\natexlab[1]{#1}
\providecommand\showeprint[2][]{arXiv:#2}

\bibitem[Alistarh et~al\mbox{.}(2015)]%
        {alistarh2015streaming}
\bibfield{author}{\bibinfo{person}{Dan Alistarh}, \bibinfo{person}{Jennifer Iglesias}, {and} \bibinfo{person}{Milan Vojnovic}.} \bibinfo{year}{2015}\natexlab{}.
\newblock \showarticletitle{Streaming min-max hypergraph partitioning}.
\newblock \bibinfo{journal}{\emph{Advances in Neural Information Processing Systems}}  \bibinfo{volume}{28} (\bibinfo{year}{2015}).
\newblock


\bibitem[Andre et~al\mbox{.}(2018)]%
        {andre2018memetic}
\bibfield{author}{\bibinfo{person}{Robin Andre}, \bibinfo{person}{Sebastian Schlag}, {and} \bibinfo{person}{Christian Schulz}.} \bibinfo{year}{2018}\natexlab{}.
\newblock \showarticletitle{Memetic multilevel hypergraph partitioning}. In \bibinfo{booktitle}{\emph{Proceedings of the Genetic and Evolutionary Computation Conference}}. \bibinfo{pages}{347--354}.
\newblock


\bibitem[Bustany et~al\mbox{.}(2023)]%
        {MLCAD2023}
\bibfield{author}{\bibinfo{person}{Ismail Bustany}, \bibinfo{person}{Grigor Gasparyan}, \bibinfo{person}{Amit Gupta}, \bibinfo{person}{Andrew~B Kahng}, \bibinfo{person}{Meghraj Kalase}, \bibinfo{person}{Wuxi Li}, {and} \bibinfo{person}{Bodhisatta Pramanik}.} \bibinfo{year}{2023}\natexlab{}.
\newblock \showarticletitle{The 2023 mlcad fpga macro placement benchmark design suite and contest results}. In \bibinfo{booktitle}{\emph{2023 ACM/IEEE 5th Workshop on Machine Learning for CAD (MLCAD)}}. IEEE, \bibinfo{pages}{1--6}.
\newblock


\bibitem[Caldwell et~al\mbox{.}(1999)]%
        {caldwell1999hypergraph}
\bibfield{author}{\bibinfo{person}{Andrew~E Caldwell}, \bibinfo{person}{Andrew~B Kahng}, \bibinfo{person}{Andrew~A Kennings}, {and} \bibinfo{person}{Igor~L Markov}.} \bibinfo{year}{1999}\natexlab{}.
\newblock \showarticletitle{Hypergraph partitioning for VLSI CAD: methodology for heuristic development, experimentation and reporting}. In \bibinfo{booktitle}{\emph{Proceedings of the 36th annual ACM/IEEE Design Automation Conference}}. \bibinfo{pages}{349--354}.
\newblock


\bibitem[{\c{C}}ataly{\"u}rek and Aykanat(2011)]%
        {ccatalyurek2011patoh}
\bibfield{author}{\bibinfo{person}{{\"U}mit {\c{C}}ataly{\"u}rek} {and} \bibinfo{person}{Cevdet Aykanat}.} \bibinfo{year}{2011}\natexlab{}.
\newblock \showarticletitle{Patoh (partitioning tool for hypergraphs)}.
\newblock In \bibinfo{booktitle}{\emph{Encyclopedia of parallel computing}}. \bibinfo{publisher}{Springer}, \bibinfo{pages}{1479--1487}.
\newblock


\bibitem[{\c{C}}ataly{\"u}rek et~al\mbox{.}(2023)]%
        {ccatalyurek2023more}
\bibfield{author}{\bibinfo{person}{{\"U}mit {\c{C}}ataly{\"u}rek}, \bibinfo{person}{Karen Devine}, \bibinfo{person}{Marcelo Faraj}, \bibinfo{person}{Lars Gottesb{\"u}ren}, \bibinfo{person}{Tobias Heuer}, \bibinfo{person}{Henning Meyerhenke}, \bibinfo{person}{Peter Sanders}, \bibinfo{person}{Sebastian Schlag}, \bibinfo{person}{Christian Schulz}, \bibinfo{person}{Daniel Seemaier}, {et~al\mbox{.}}} \bibinfo{year}{2023}\natexlab{}.
\newblock \showarticletitle{More recent advances in (hyper) graph partitioning}.
\newblock \bibinfo{journal}{\emph{Comput. Surveys}} \bibinfo{volume}{55}, \bibinfo{number}{12} (\bibinfo{year}{2023}), \bibinfo{pages}{1--38}.
\newblock


\bibitem[Catalyurek and Aykanat(2002)]%
        {catalyurek2002hypergraph}
\bibfield{author}{\bibinfo{person}{Umit~V Catalyurek} {and} \bibinfo{person}{Cevdet Aykanat}.} \bibinfo{year}{2002}\natexlab{}.
\newblock \showarticletitle{Hypergraph-partitioning-based decomposition for parallel sparse-matrix vector multiplication}.
\newblock \bibinfo{journal}{\emph{IEEE Transactions on parallel and distributed systems}} \bibinfo{volume}{10}, \bibinfo{number}{7} (\bibinfo{year}{2002}), \bibinfo{pages}{673--693}.
\newblock


\bibitem[Devine et~al\mbox{.}(2006)]%
        {devine2006parallel}
\bibfield{author}{\bibinfo{person}{Karen~D Devine}, \bibinfo{person}{Erik~G Boman}, \bibinfo{person}{Robert~T Heaphy}, \bibinfo{person}{Rob~H Bisseling}, {and} \bibinfo{person}{Umit~V Catalyurek}.} \bibinfo{year}{2006}\natexlab{}.
\newblock \showarticletitle{Parallel hypergraph partitioning for scientific computing}. In \bibinfo{booktitle}{\emph{Proceedings 20th IEEE International Parallel \& Distributed Processing Symposium}}. IEEE, \bibinfo{pages}{10--pp}.
\newblock


\bibitem[Duraira and Kalla(2004)]%
        {duraira2004exploiting}
\bibfield{author}{\bibinfo{person}{V Duraira} {and} \bibinfo{person}{Priyank Kalla}.} \bibinfo{year}{2004}\natexlab{}.
\newblock \showarticletitle{Exploiting hypergraph partitioning for efficient boolean satisfiability}. In \bibinfo{booktitle}{\emph{Proceedings. Ninth IEEE International High-Level Design Validation and Test Workshop (IEEE Cat. No. 04EX940)}}. IEEE, \bibinfo{pages}{141--146}.
\newblock


\bibitem[ERDdS and R\&wi(1959)]%
        {erdds1959random}
\bibfield{author}{\bibinfo{person}{P ERDdS} {and} \bibinfo{person}{A R\&wi}.} \bibinfo{year}{1959}\natexlab{}.
\newblock \showarticletitle{On random graphs I}.
\newblock \bibinfo{journal}{\emph{Publ. math. debrecen}} \bibinfo{volume}{6}, \bibinfo{number}{290-297} (\bibinfo{year}{1959}), \bibinfo{pages}{18}.
\newblock


\bibitem[Eyubov et~al\mbox{.}(2025)]%
        {eyubov2025freight}
\bibfield{author}{\bibinfo{person}{Kamal Eyubov}, \bibinfo{person}{Marcelo Fonseca~Faraj}, {and} \bibinfo{person}{Christian Schulz}.} \bibinfo{year}{2025}\natexlab{}.
\newblock \showarticletitle{FREIGHT: Fast streaming hypergraph partitioning}.
\newblock \bibinfo{journal}{\emph{Algorithmica}} \bibinfo{volume}{87}, \bibinfo{number}{3} (\bibinfo{year}{2025}), \bibinfo{pages}{405--428}.
\newblock


\bibitem[Faraj and Schulz(2022)]%
        {faraj2022buffered}
\bibfield{author}{\bibinfo{person}{Marcelo~Fonseca Faraj} {and} \bibinfo{person}{Christian Schulz}.} \bibinfo{year}{2022}\natexlab{}.
\newblock \showarticletitle{Buffered streaming graph partitioning}.
\newblock \bibinfo{journal}{\emph{ACM Journal of Experimental Algorithmics}}  \bibinfo{volume}{27} (\bibinfo{year}{2022}), \bibinfo{pages}{1--26}.
\newblock


\bibitem[Gao et~al\mbox{.}(2020)]%
        {gao2020hypergraph}
\bibfield{author}{\bibinfo{person}{Yue Gao}, \bibinfo{person}{Zizhao Zhang}, \bibinfo{person}{Haojie Lin}, \bibinfo{person}{Xibin Zhao}, \bibinfo{person}{Shaoyi Du}, {and} \bibinfo{person}{Changqing Zou}.} \bibinfo{year}{2020}\natexlab{}.
\newblock \showarticletitle{Hypergraph learning: Methods and practices}.
\newblock \bibinfo{journal}{\emph{IEEE transactions on pattern analysis and machine intelligence}} \bibinfo{volume}{44}, \bibinfo{number}{5} (\bibinfo{year}{2020}), \bibinfo{pages}{2548--2566}.
\newblock


\bibitem[Gottesb{\"u}ren et~al\mbox{.}(2024)]%
        {gottesburen2024scalable}
\bibfield{author}{\bibinfo{person}{Lars Gottesb{\"u}ren}, \bibinfo{person}{Tobias Heuer}, \bibinfo{person}{Nikolai Maas}, \bibinfo{person}{Peter Sanders}, {and} \bibinfo{person}{Sebastian Schlag}.} \bibinfo{year}{2024}\natexlab{}.
\newblock \showarticletitle{Scalable high-quality hypergraph partitioning}.
\newblock \bibinfo{journal}{\emph{ACM Transactions on Algorithms}} \bibinfo{volume}{20}, \bibinfo{number}{1} (\bibinfo{year}{2024}), \bibinfo{pages}{1--54}.
\newblock


\bibitem[Gottesb{\"u}ren et~al\mbox{.}(2022)]%
        {gottesburen2022shared}
\bibfield{author}{\bibinfo{person}{Lars Gottesb{\"u}ren}, \bibinfo{person}{Tobias Heuer}, \bibinfo{person}{Peter Sanders}, {and} \bibinfo{person}{Sebastian Schlag}.} \bibinfo{year}{2022}\natexlab{}.
\newblock \showarticletitle{Shared-Memory n-level Hypergraph Partitioning}. In \bibinfo{booktitle}{\emph{2022 Proceedings of the Symposium on Algorithm Engineering and Experiments (ALENEX)}}. SIAM, \bibinfo{pages}{131--144}.
\newblock


\bibitem[Guo et~al\mbox{.}(2023)]%
        {guo2023lahypergraph}
\bibfield{author}{\bibinfo{person}{Luanzheng Guo}, \bibinfo{person}{Jesun Firoz}, {and} \bibinfo{person}{Gokcen Kestor}.} \bibinfo{year}{2023}\natexlab{}.
\newblock \showarticletitle{LAHypergraph: Parallel Hypergraph Analytics in the Language of Linear Algebra}. In \bibinfo{booktitle}{\emph{SIAM Conference on Applied and Computational Discrete Algorithms (ACDA23)}}. SIAM, \bibinfo{pages}{147--158}.
\newblock


\bibitem[Jiang(2018)]%
        {jiang2018highly}
\bibfield{author}{\bibinfo{person}{Wenkai Jiang}.} \bibinfo{year}{2018}\natexlab{}.
\newblock \showarticletitle{Highly efficient distributed hypergraph analysis: Real-time partitioning and quantized learning}.
\newblock \bibinfo{journal}{\emph{Master's thesis, Univ. Melbourne}} (\bibinfo{year}{2018}).
\newblock


\bibitem[Kami{\'n}ski et~al\mbox{.}(2019)]%
        {kaminski2019clustering}
\bibfield{author}{\bibinfo{person}{Bogumi{\l} Kami{\'n}ski}, \bibinfo{person}{Val{\'e}rie Poulin}, \bibinfo{person}{Pawe{\l} Pra{\l}at}, \bibinfo{person}{Przemys{\l}aw Szufel}, {and} \bibinfo{person}{Fran{\c{c}}ois Th{\'e}berge}.} \bibinfo{year}{2019}\natexlab{}.
\newblock \showarticletitle{Clustering via hypergraph modularity}.
\newblock \bibinfo{journal}{\emph{PloS one}} \bibinfo{volume}{14}, \bibinfo{number}{11} (\bibinfo{year}{2019}), \bibinfo{pages}{e0224307}.
\newblock


\bibitem[Karypis et~al\mbox{.}(1997)]%
        {karypis1997multilevel}
\bibfield{author}{\bibinfo{person}{George Karypis}, \bibinfo{person}{Rajat Aggarwal}, \bibinfo{person}{Vipin Kumar}, {and} \bibinfo{person}{Shashi Shekhar}.} \bibinfo{year}{1997}\natexlab{}.
\newblock \showarticletitle{Multilevel hypergraph partitioning: Application in VLSI domain}. In \bibinfo{booktitle}{\emph{Proceedings of the 34th annual Design Automation Conference}}. \bibinfo{pages}{526--529}.
\newblock


\bibitem[Karypis and Kumar(1999)]%
        {karypis1999multilevel}
\bibfield{author}{\bibinfo{person}{George Karypis} {and} \bibinfo{person}{Vipin Kumar}.} \bibinfo{year}{1999}\natexlab{}.
\newblock \showarticletitle{Multilevel k-way hypergraph partitioning}. In \bibinfo{booktitle}{\emph{Proceedings of the 36th annual ACM/IEEE design automation conference}}. \bibinfo{pages}{343--348}.
\newblock


\bibitem[Lee et~al\mbox{.}(2025)]%
        {lee2025survey}
\bibfield{author}{\bibinfo{person}{Geon Lee}, \bibinfo{person}{Fanchen Bu}, \bibinfo{person}{Tina Eliassi-Rad}, {and} \bibinfo{person}{Kijung Shin}.} \bibinfo{year}{2025}\natexlab{}.
\newblock \showarticletitle{A survey on hypergraph mining: Patterns, tools, and generators}.
\newblock \bibinfo{journal}{\emph{Comput. Surveys}} \bibinfo{volume}{57}, \bibinfo{number}{8} (\bibinfo{year}{2025}), \bibinfo{pages}{1--36}.
\newblock


\bibitem[Liang et~al\mbox{.}(2024)]%
        {liang2024medpart}
\bibfield{author}{\bibinfo{person}{Rongjian Liang}, \bibinfo{person}{Anthony Agnesina}, {and} \bibinfo{person}{Haoxing Ren}.} \bibinfo{year}{2024}\natexlab{}.
\newblock \showarticletitle{Medpart: A multi-level evolutionary differentiable hypergraph partitioner}. In \bibinfo{booktitle}{\emph{Proceedings of the 2024 International Symposium on Physical Design}}. \bibinfo{pages}{3--11}.
\newblock


\bibitem[Markov et~al\mbox{.}(2012)]%
        {markov2012progress}
\bibfield{author}{\bibinfo{person}{Igor~L Markov}, \bibinfo{person}{Jin Hu}, {and} \bibinfo{person}{Myung-Chul Kim}.} \bibinfo{year}{2012}\natexlab{}.
\newblock \showarticletitle{Progress and challenges in VLSI placement research}. In \bibinfo{booktitle}{\emph{Proceedings of the International Conference on Computer-Aided Design}}. \bibinfo{pages}{275--282}.
\newblock


\bibitem[Mayer et~al\mbox{.}(2018)]%
        {mayer2018hype}
\bibfield{author}{\bibinfo{person}{Christian Mayer}, \bibinfo{person}{Ruben Mayer}, \bibinfo{person}{Sukanya Bhowmik}, \bibinfo{person}{Lukas Epple}, {and} \bibinfo{person}{Kurt Rothermel}.} \bibinfo{year}{2018}\natexlab{}.
\newblock \showarticletitle{HYPE: massive hypergraph partitioning with neighborhood expansion}. In \bibinfo{booktitle}{\emph{2018 IEEE International Conference on Big Data (Big Data)}}. IEEE, \bibinfo{pages}{458--467}.
\newblock


\bibitem[McGregor(2014)]%
        {mcgregor2014graph}
\bibfield{author}{\bibinfo{person}{Andrew McGregor}.} \bibinfo{year}{2014}\natexlab{}.
\newblock \showarticletitle{Graph stream algorithms: a survey}.
\newblock \bibinfo{journal}{\emph{ACM SIGMOD Record}} \bibinfo{volume}{43}, \bibinfo{number}{1} (\bibinfo{year}{2014}), \bibinfo{pages}{9--20}.
\newblock


\bibitem[Nam(2006)]%
        {ISPD2006}
\bibfield{author}{\bibinfo{person}{Gi-Joon Nam}.} \bibinfo{year}{2006}\natexlab{}.
\newblock \showarticletitle{ISPD 2006 Placement Contest: Benchmark Suite and Results}. In \bibinfo{booktitle}{\emph{Proceedings of the 2006 International Symposium on Physical Design}} (San Jose, California, USA) \emph{(\bibinfo{series}{ISPD '06})}. \bibinfo{publisher}{Association for Computing Machinery}, \bibinfo{address}{New York, NY, USA}, \bibinfo{pages}{167}.
\newblock
\showISBNx{1595932992}
\href{https://doi.org/10.1145/1123008.1123042}{doi:\nolinkurl{10.1145/1123008.1123042}}


\bibitem[Papa and Markov(2007)]%
        {papa2007hypergraph}
\bibfield{author}{\bibinfo{person}{David~A Papa} {and} \bibinfo{person}{Igor~L Markov}.} \bibinfo{year}{2007}\natexlab{}.
\newblock \showarticletitle{Hypergraph Partitioning and Clustering.}
\newblock \bibinfo{journal}{\emph{Handbook of Approximation Algorithms and Metaheuristics}}  \bibinfo{volume}{20073547} (\bibinfo{year}{2007}), \bibinfo{pages}{61--1}.
\newblock


\bibitem[Patwary et~al\mbox{.}(2019)]%
        {patwary2019window}
\bibfield{author}{\bibinfo{person}{Md~Anwarul~Kaium Patwary}, \bibinfo{person}{Saurabh Garg}, {and} \bibinfo{person}{Byeong Kang}.} \bibinfo{year}{2019}\natexlab{}.
\newblock \showarticletitle{Window-based streaming graph partitioning algorithm}. In \bibinfo{booktitle}{\emph{Proceedings of the australasian computer science week multiconference}}. \bibinfo{pages}{1--10}.
\newblock


\bibitem[Popp et~al\mbox{.}(2021)]%
        {popp2021multilevel}
\bibfield{author}{\bibinfo{person}{Merten Popp}, \bibinfo{person}{Sebastian Schlag}, \bibinfo{person}{Christian Schulz}, {and} \bibinfo{person}{Daniel Seemaier}.} \bibinfo{year}{2021}\natexlab{}.
\newblock \showarticletitle{Multilevel Acyclic Hypergraph Partitioning}. In \bibinfo{booktitle}{\emph{2021 Proceedings of the Workshop on Algorithm Engineering and Experiments (ALENEX)}}. SIAM, \bibinfo{pages}{1--15}.
\newblock


\bibitem[Schlag et~al\mbox{.}(2023)]%
        {schlag2023high}
\bibfield{author}{\bibinfo{person}{Sebastian Schlag}, \bibinfo{person}{Tobias Heuer}, \bibinfo{person}{Lars Gottesb{\"u}ren}, \bibinfo{person}{Yaroslav Akhremtsev}, \bibinfo{person}{Christian Schulz}, {and} \bibinfo{person}{Peter Sanders}.} \bibinfo{year}{2023}\natexlab{}.
\newblock \showarticletitle{High-quality hypergraph partitioning}.
\newblock \bibinfo{journal}{\emph{ACM Journal of Experimental Algorithmics}}  \bibinfo{volume}{27} (\bibinfo{year}{2023}), \bibinfo{pages}{1--39}.
\newblock


\bibitem[Stanton(2014)]%
        {stanton2014streaming}
\bibfield{author}{\bibinfo{person}{Isabelle Stanton}.} \bibinfo{year}{2014}\natexlab{}.
\newblock \showarticletitle{Streaming balanced graph partitioning algorithms for random graphs}. In \bibinfo{booktitle}{\emph{Proceedings of the twenty-fifth annual ACM-SIAM symposium on Discrete algorithms}}. SIAM, \bibinfo{pages}{1287--1301}.
\newblock


\bibitem[Sun et~al\mbox{.}(2008)]%
        {sun2008hypergraph}
\bibfield{author}{\bibinfo{person}{Liang Sun}, \bibinfo{person}{Shuiwang Ji}, {and} \bibinfo{person}{Jieping Ye}.} \bibinfo{year}{2008}\natexlab{}.
\newblock \showarticletitle{Hypergraph spectral learning for multi-label classification}. In \bibinfo{booktitle}{\emph{Proceedings of the 14th ACM SIGKDD international conference on Knowledge discovery and data mining}}. \bibinfo{pages}{668--676}.
\newblock


\bibitem[Ta{\c{s}}yaran et~al\mbox{.}(2021)]%
        {tacsyaran2021streaming}
\bibfield{author}{\bibinfo{person}{Fatih Ta{\c{s}}yaran}, \bibinfo{person}{Berkay Demireller}, \bibinfo{person}{Kamer Kaya}, {and} \bibinfo{person}{Bora U{\c{c}}ar}.} \bibinfo{year}{2021}\natexlab{}.
\newblock \showarticletitle{Streaming Hypergraph Partitioning Algorithms on Limited Memory Environments}. In \bibinfo{booktitle}{\emph{HPCS 2020-International Conference on High Performance Computing \& Simulation}}. IEEE, \bibinfo{pages}{1--8}.
\newblock


\bibitem[Tsourakakis et~al\mbox{.}(2014)]%
        {tsourakakis2014fennel}
\bibfield{author}{\bibinfo{person}{Charalampos Tsourakakis}, \bibinfo{person}{Christos Gkantsidis}, \bibinfo{person}{Bozidar Radunovic}, {and} \bibinfo{person}{Milan Vojnovic}.} \bibinfo{year}{2014}\natexlab{}.
\newblock \showarticletitle{Fennel: Streaming graph partitioning for massive scale graphs}. In \bibinfo{booktitle}{\emph{Proceedings of the 7th ACM international conference on Web search and data mining}}. \bibinfo{pages}{333--342}.
\newblock


\bibitem[Yang et~al\mbox{.}(2022)]%
        {yang2022semi}
\bibfield{author}{\bibinfo{person}{Chaoqi Yang}, \bibinfo{person}{Ruijie Wang}, \bibinfo{person}{Shuochao Yao}, {and} \bibinfo{person}{Tarek Abdelzaher}.} \bibinfo{year}{2022}\natexlab{}.
\newblock \showarticletitle{Semi-supervised hypergraph node classification on hypergraph line expansion}. In \bibinfo{booktitle}{\emph{Proceedings of the 31st ACM international conference on information \& knowledge management}}. \bibinfo{pages}{2352--2361}.
\newblock


\bibitem[Zien et~al\mbox{.}(2002)]%
        {zien2002multilevel}
\bibfield{author}{\bibinfo{person}{Jason~Y Zien}, \bibinfo{person}{Martine~DF Schlag}, {and} \bibinfo{person}{Pak~K Chan}.} \bibinfo{year}{2002}\natexlab{}.
\newblock \showarticletitle{Multilevel spectral hypergraph partitioning with arbitrary vertex sizes}.
\newblock \bibinfo{journal}{\emph{IEEE Transactions on computer-aided design of integrated circuits and systems}} \bibinfo{volume}{18}, \bibinfo{number}{9} (\bibinfo{year}{2002}), \bibinfo{pages}{1389--1399}.
\newblock


\end{thebibliography}
\end{document}